\documentclass[runningheads]{llncs}
\usepackage{listings}
\usepackage{subcaption}
\usepackage{graphicx}
\usepackage{todonotes}
\usepackage{hyperref}
\usepackage{url}
\usepackage{multirow}
\usepackage{tabularx}
\usepackage{ltablex}
\usepackage{glossaries}
\usepackage[strings]{underscore}
\usepackage[english]{babel}
\addto\extrasenglish{

}

\newcommand{\mlflow}{MLFlow }
\newcommand{\mlflows}{MLFlow's }
\newcommand{\spacy}{SpaCy }
\newcommand{\alstratpredot}{AL strategy}

\newcommand{\alstrat}{AL strategy }
\newcommand{\alstrats}{AL strategies }

\definecolor{codegreen}{rgb}{0,0.6,0}
\definecolor{codegray}{rgb}{0.5,0.5,0.5}
\definecolor{codepurple}{rgb}{0.58,0,0.82}
\definecolor{backcolour}{rgb}{0.95,0.95,0.92}

\lstdefinestyle{python}{
    language=Python,
    backgroundcolor=\color{backcolour},   
    commentstyle=\color{codegreen},
    keywordstyle=\color{magenta},
    numberstyle=\tiny\color{codegray},
    stringstyle=\color{codepurple},
    basicstyle={\tiny\ttfamily},
    breakatwhitespace=false,         
    breaklines=true,                 
    captionpos=b,                    
    keepspaces=true,                 
    numbers=left,                    
    numbersep=5pt,                  
    showspaces=false,                
    showstringspaces=false,
    showtabs=false,                  
    tabsize=2
}

\begin{document}

\title{ALE: A Simulation-Based Active Learning Evaluation Framework for the Parameter-Driven Comparison of Query Strategies for NLP}
\titlerunning{Active Learning Evaluation (ALE) Framework}
%
\author{Philipp Kohl\inst{1}\orcidID{0000-0002-5972-8413} \and Nils Freyer\inst{1}\orcidID{0000-0002-4460-3650}
\and Yoka Krämer\inst{1}\orcidID{0009-0006-7326-3268} \and Henri Werth\inst{3}\orcidID{0009-0001-6626-3159} \and Steffen Wolf\inst{1}\orcidID{0009-0005-7987-5479} \and Bodo Kraft\inst{1}\and Matthias Meinecke\inst{1}\orcidID{0009-0008-3055-5505} \and Albert Zündorf\inst{2}}
\authorrunning{Kohl et al.}
\institute{Anonymous}
\institute{FH Aachen -- University of Applied Sciences, 52428 Jülich, Germany \email{\{p.kohl,freyer,y.kraemer,s.wolf,kraft,meinecke\}@fh-aachen.de}\and
University of Kassel, 34121 Kassel, Germany
\email{zuendorf@uni-kassel.de}\\ 
Laizee.ai \email{henri@laizee.ai}\\ \url{https://laizee.ai/}}
\maketitle              
\begin{abstract}
Supervised machine learning and deep learning require a large amount of labeled data, which data scientists obtain in a manual, and time-consuming annotation process. To mitigate this challenge, Active Learning (AL) proposes promising data points to annotators they annotate next instead of a subsequent or random sample. This method is supposed to save annotation effort while maintaining model performance.

However, practitioners face many \alstrats for different tasks and need an empirical basis to choose between them. Surveys categorize \alstrats into taxonomies without performance indications. Presentations of novel \alstrats compare the performance to a small subset of strategies.
Our contribution addresses the empirical basis by introducing a reproducible active learning evaluation (ALE) framework for the comparative evaluation of \alstrats in NLP. 

The framework allows the implementation of \alstrats with low effort and a fair data-driven comparison through defining and tracking experiment parameters (e.g., initial dataset size, number of data points per query step,  and the budget). ALE helps practitioners to make more informed decisions, and researchers can focus on developing new, effective \alstrats and deriving best practices for specific use cases. With best practices, practitioners can lower their annotation costs. We present a case study to illustrate how to use the framework.

\end{abstract}

\section{Introduction}
\label{sec:introduction}
Within the last decades, machine learning (ML) and, more specifically, deep learning (DL) have opened up great potential in natural language processing (NLP) \cite{devlinBERTPretrainingDeep2019,brownLanguageModelsAre2020}.
Supervised machine learning algorithms constitute a powerful class of algorithms for NLP tasks like text classification, named entity recognition (NER), and relation extraction \cite{nasarNamedEntityRecognition2022,kowsariTextClassificationAlgorithms2019}. Researchers and corporations require large amounts of annotated data to develop state-of-the-art applications. However, domain-specific labeled data is often scarce, and the annotation process is time, and cost-intensive \cite{herdeSurveyCostTypes2021}. Therefore, reducing the amount of data needed to train a model is an active field of research \cite{herdeSurveyCostTypes2021}. Practitioners and researchers agree on the need to reduce annotation costs. Estimating these expenses is a research field on its own \cite{herdeSurveyCostTypes2021,aroraEstimatingAnnotationCost2009,tomanekApproachTextCorpus}. 

\begin{figure}[htb]
    \centering
    \includegraphics[width=0.85\textwidth]{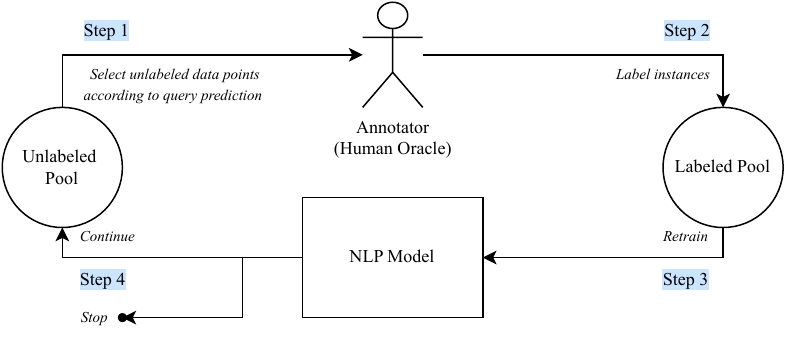}
    \caption[Active learning cycle (following \cite{settlesActiveLearning2012,herdeSurveyCostTypes2021})]{Active learning cycle (following \cite{settlesActiveLearning2012,herdeSurveyCostTypes2021}): \textbf{1st step}: The AL strategy proposes the next batch of data points to the annotator. \textbf{2nd step}: The annotator labels the given increment. \textbf{3rd step}: The training component trains a model using the updated set of labeled data points. \textbf{4th step}: Stop the process if the model achieved the desired performance, or the unlabeled pool is empty. Otherwise, repeat the process.}
    \label{fig:ALCycle}
\end{figure}

Next to transfer learning \cite{ruderTransferLearningNatural} from large language models or few-shot learning \cite{songComprehensiveSurveyFewshot2022}, active learning (AL) is another approach to reduce the required data size by systematically annotating data \cite{sunSurveyActiveLearning2010,schroderSurveyActiveLearning2020}.
\alstrats (also called \textit{query} or \textit{teacher}) propose data points to the annotator in the annotation process, which could increase the model's performance more than a randomly chosen data point.
Consequently, the model may perform better with fewer data when annotating the proposed data points instead of considering random ones.
\autoref{fig:ALCycle} shows the human-in-the-loop paradigm with pool-based annotation approaches. Instead of selecting random or sequential data points, as processed in a standard annotation process, \alstrats forward beneficial samples to the annotator.
Active learning has empirically proven significant advances in specific domains,
including computer vision or activity recognition \cite{alemdarhandeActiveLearningUncertainty2017,loyStreambasedJointExplorationexploitation2012}. \cite{fengDeepActiveLearning2019,duBreastCancerHistopathological2018} report up to 60\% of annotation effort reduction.

The performance of an \alstrat is typically measured by the performance metric of the corresponding model when trained on the same amount of systematically proposed data compared to randomly proposed data.  
However, \alstrats often use heuristics that may be task, data, or model sensitive \cite{schroderSurveyActiveLearning2020}. 
Little can be said about their performance in a comparable way. 
Surveys often classify \alstrats into different categories but do not provide exhaustive performance benchmarks \cite{renSurveyDeepActive2022,settlesActiveLearning2012}. 
Researchers developing new \alstrats compare their performance with random sampling and a few other strategies \cite{guCombiningActiveLearning2014,yanClusteringbasedActiveLearning2022}.
The research field of NLP lacks a standardized procedure to evaluate \alstrats systematically for a given task, dataset, and model.

ALE facilitates researchers and practitioners to gain more insights into task-dependent taxonomies and best practices. The goal is to enable practitioners to adapt the most valuable \alstrat based on comparable previous experiments.
Therefore, this paper introduces the Active Learning Evaluation (ALE)\footnote{\url{https://github.com/philipp-kohl/Active-Learning-Evaluation-Framework}} Framework.
Our framework aims to close the \textit{comparison gap} \cite{zhanComparativeSurveyDeep2022} between different AL strategies.
ALE enables researchers in NLP to implement and evaluate their own \alstrats in a reproducible and transparent way. By offering default implementations, practitioners can directly compare their \alstrats and datasets. Using the \textit{\mlflow} platform\footnote{\url{https://mlflow.org/}}, ALE simplifies visualizing and reporting experimental results.
At the same time, practitioners can customize the configurations and adapt the framework to their specific requirements. 

\section{Related Work}
\label{sec:related-work}
Active learning started with classical machine learning tasks (such as support vector machines, decision trees, etc.) \cite{settlesActiveLearning2012}. With the rise of deep learning architectures, researchers transfer the knowledge of classical approaches to those using deep learning \cite{renSurveyDeepActive2022}. Active learning has gained significant attention across various domains, such as computer vision and natural language processing (NLP) \cite{renSurveyDeepActive2022,zhanComparativeSurveyDeep2022}.
In computer vision, users apply active learning to tasks like image classification, object detection, and video processing. In NLP, it has been used for text classification, entity recognition, and machine translation. 

Researchers classify active learning approaches based on several taxonomies:
(1) Explorative and exploitative (\cite{bonduExplorationVsExploitation2010,loyStreambasedJointExplorationexploitation2012}) approaches: Exploration is model-agnostic and uses information about the data distribution, while exploitation uses model information to propose valuable data points.   
(2) \cite{settlesActiveLearning2012} distinguish stream-based from pool-based approaches. Stream-based methods sample a single data point and decide whether to present it to the human oracle (annotator). In contrast, the pool-based method selects a subset of the unlabeled dataset.
(3) \cite{herdeSurveyCostTypes2021} compares traditional and real-world active learning, which pays attention to the complexity of the actual annotation process. Real-world strategies do not assume a perfect oracle but take into account that humans are individuals that make mistakes. Relying on a single human annotation as  the gold label can result in errors. Addressing this issue impacts the annotation time, cost, and quality.

Despite using a single AL strategy, \cite{mendoncaQueryStrategiesAssemble2020} shows how to determine the most valuable method from a pre-selected subset on the fly. The authors select the best strategy by incrementally re-weighting the number of new data points each strategy proposes while performing the active learning cycle (see \autoref{fig:ALCycle}). Thus, the method finds the subset's best strategy over iterations. ALE does not address the dynamic selection of the best strategy in the annotation process, but it helps to compare the sub-selected \alstrats in isolation to select a promising strategy for the annotation process. But understanding \cite{mendoncaQueryStrategiesAssemble2020}'s approach as a single AL strategy, ALE can compare this method to others.

Researchers developed many active learning strategies in various domains \cite{settlesActiveLearning2012,renSurveyDeepActive2022}. However, annotation tools for NLP tasks seem to prefer the basic method of uncertainty sampling as AL strategy:
INCEpTION \cite{klieINCEpTIONPlatformMachineAssisted2018} or Prodigy \cite{montaniProdigyModernScriptable}. Other tools like doccano \cite{nakayamaDoccanoTextAnnotation2018} plan to implement active learning in the future. 

The taxonomies create a significant challenge for practitioners 
seeking to adopt active learning for their specific use case. Developing best practices for similar use cases is necessary to address this challenge.
We created an evaluation framework for \alstrats in a consistent manner. This framework allows practitioners to compare different AL strategies using the same set of parameters, thereby facilitating the selection of the most appropriate strategy for their specific use case. Surveys on active learning propose various methodologies, but they are limited in their ability to provide a comprehensive performance comparison of different \alstrats\cite{sunSurveyActiveLearning2010,settlesActiveLearning2012,renSurveyDeepActive2022}. \cite{zhanComparativeSurveyDeep2022} provides a performance comparison for 19 AL strategies. 

Similar to our objective, \cite{yangLibactPoolbasedActive2017} implemented \textit{libact}, an active learning comparison and application framework for classical machine learning methods with scikit-learn \cite{pedregosaScikitlearnMachineLearning2011}. \textit{libact} offers the feature to automatically select the best-performing strategy by the \textit{Active Learning By Learning (ALBL)} algorithm \cite{hsuActiveLearningLearning2015} over iterations in the annotation process.
While \textit{libact} focuses on classical machine learning with scitkit-learn, ALE enables users to apply their preferred ML/DL tools, such as scikit-learn, TensorFlow, PyTorch, or SpaCy. We want to inspect the NLP domain and choose the deep learning framework \textit{SpaCy}\footnote{\url{https://spacy.io/}} as the default implementation.

\cite{huangDeepALDeepActive2021,zhanComparativeSurveyDeep2022} developed a CLI tool (\textit{DeepAL}) to compare different pool-based \alstrats with PyTorch\footnote{\url{https://pytorch.org/}} models in the image vision domain. 
ALE focuses on the NLP domain and offers a sophisticated experiment configuration and tracking management. This leads to a reproducible and transparent experiment environment. Furthermore, we use parallelization and dockerization
to address the long computation times and enable cloud infrastructure usage. Additionally, we facilitate experiments on the cold start phase \cite{yuanColdstartActiveLearning2020} with the same strategies as for active learning.

\section{Ethical Considerations}
\label{sec:ethical-considerations}
Despite the ethical considerations regarding NLP \cite{leidnerEthicalDesignEthics2017b,weidingerEthicalSocialRisks,benderDangersStochasticParrots2021}, more specific ethical implications are worth evaluating for active learning, as it may introduce or reinforce statistical biases to the data \cite{farquharStatisticalBiasActive2021}. Especially regarding safety or life path-relevant decision-making, introducing biases to the training data may cause ethically significant harms \cite{vallorTechnologyVirtuesPhilosophical2016}. Thus, when applying active learning strategies, it is urgent to monitor and control biases in the resulting data\footnote{A similar concern was made within the \href{https://digital-strategy.ec.europa.eu/en/policies/regulatory-framework-ai}{regulatory framework proposal on artificial intelligence} by the European Union, classifying the respective systems as \textit{high risk}.}.
In contrast, efforts have been made to diminish biases in already biased datasets using \alstrats that optimize for certain fairness metrics 
\cite{anahidehFairActiveLearning2021}. Further research in that direction would be desirable and is generally supported by the ALE framework by adding additional metrics such as statistical parity \cite{anahidehFairActiveLearning2021}.

\section{ALE Framework}
\label{sec:ale-framework}
\begin{figure}[htb]
    \centering
    \includegraphics[width=0.7\linewidth]{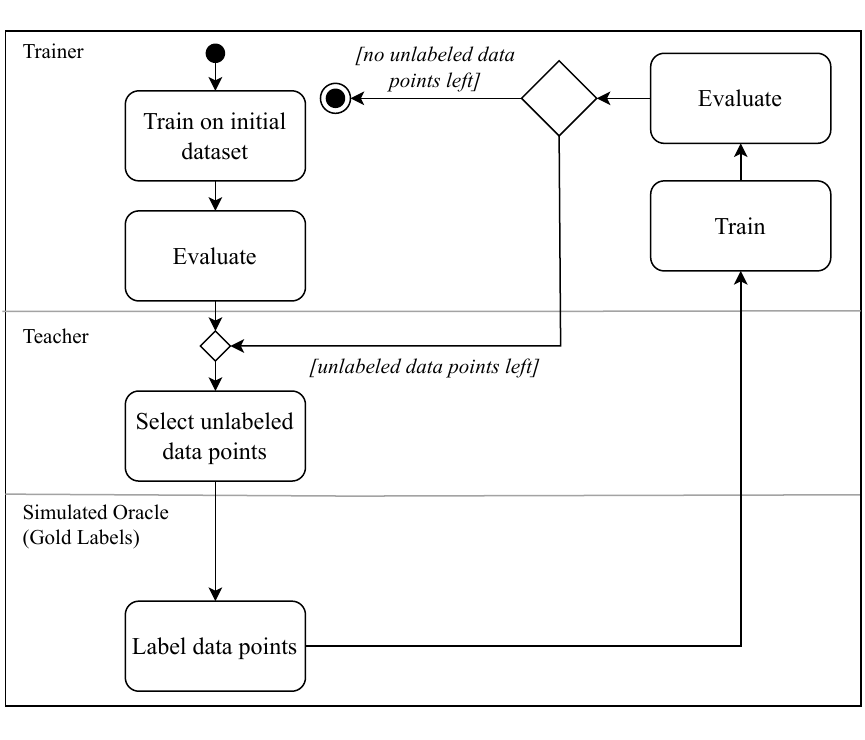}
    \caption[The simulated AL cycle realized by the ALE framework]{The \textit{simulator} manages the AL simulation cycle: The trainer starts the initial training on the first increment of data points proposed by a teacher component (omitted in this figure) and evaluates it. The shown teacher proposes unlabeled data points to the simulated oracle, which simulates the human annotator by adding the according gold label. The updated dataset is used for another training and evaluation cycle. Once there are no unlabeled data points left, the process stops.}
    \label{fig:SimulationCycle}
\end{figure}

We want to facilitate researchers and practitioners to draw data-based decisions for selecting an appropriate \alstrat for their use case.
To assess the AL strategy's performance, we simulate the human-in-the-loop as a perfect annotating oracle using the gold label from the used dataset (see \autoref{fig:SimulationCycle}). After each query step, we train a model and evaluate its performance on the test set.
These metrics serve as a performance indicator of the AL strategy. The random selection of data points represents the baseline strategy. We average the performance across several random seeds to address
the model's and strategy's stability \cite{phamProblemsOpportunitiesTraining2021,madhyasthaModelStabilityFunction2019}. 

In the following sections, we describe the features and implementation of the ALE framework in detail:

\textbf{Configuration System.} To address varying use cases, we have created a sophisticated configuration system. The user controls the parameters for the active learning simulation: initial dataset size, number of proposed data points by query, comparison metric, budget, and dataset.\footnote{Full configuration documentation can be found on GitHub.}

\textbf{Experiment documentation and tracking.} The framework documents each experiment with the configuration and results (e.g., metrics and trained models) of all substeps in MLFlow. The tracking facilitates own subsequent analysis to reveal regularities.

\textbf{Resume experiments.} Active learning simulation computation is time-consuming (hours to days), depending on the configuration (especially dataset size and step size). The training processes determine the majority of the consumption part. Thus, the framework can continue experiments on pauses or errors.

\textbf{Reproducible.} In conjunction with the experiment configuration, including fixed seeds and persisting the git commit hash enables reproducible research. 

\textbf{Average with different seeds.} To avoid beneficial or harmful seeds, the user can specify more than one seed. The framework processes different simulation runs with these seeds and averages the results. Thus, we achieve a less random biased result \cite{phamProblemsOpportunitiesTraining2021,madhyasthaModelStabilityFunction2019}. If the results show high variances, the model or the query strategy might not be stable. We avoid simulating the active learning process for each seed sequentially. Instead, we offer a configurable amount of concurrent threads to simulate the process.

\textbf{Arbitrary usage of datasets.} The user can employ their preferred datasets. No additional code must be written as long as the dataset aligns with the conventions. Otherwise, the user has to implement a new converter class. See details in \autoref{sec:dataset}.

\textbf{Arbitrary usage of ML/DL framework.} The framework uses \spacy as the default deep learning framework. The user can change the framework by implementing the trainer (\autoref{sec:trainer}) and corpus abstraction layer.

\textbf{Easy to test own strategies.} The users can implement their own configurable strategies by adding a config file, implementing a single class, and registering the class with the 
\textit{TeacherRegistry}. See the usage in the case study in \autoref{sec:case-study}.

\textbf{Experiment with cold start.} Analog to AL strategies, the user can test different strategies for the initial training dataset. Initial training strategies and \alstrats can be used interchangeably.

\textbf{Containerized.} We provide all involved services as docker services. In consequence, the environment can be bootstrapped rapidly. Furthermore, we provide a docker image for the framework facilitating the run of experiments on cloud infrastructure.

\textbf{Parallel computation.} In conjunction with the containerization, the user can run experiments in parallel, and the framework reports the result to a central \mlflow instance.

\section{Architecture}
\label{sec:architecture}

\begin{figure}[htb]
    \centering
    \includegraphics[width=0.9\linewidth]{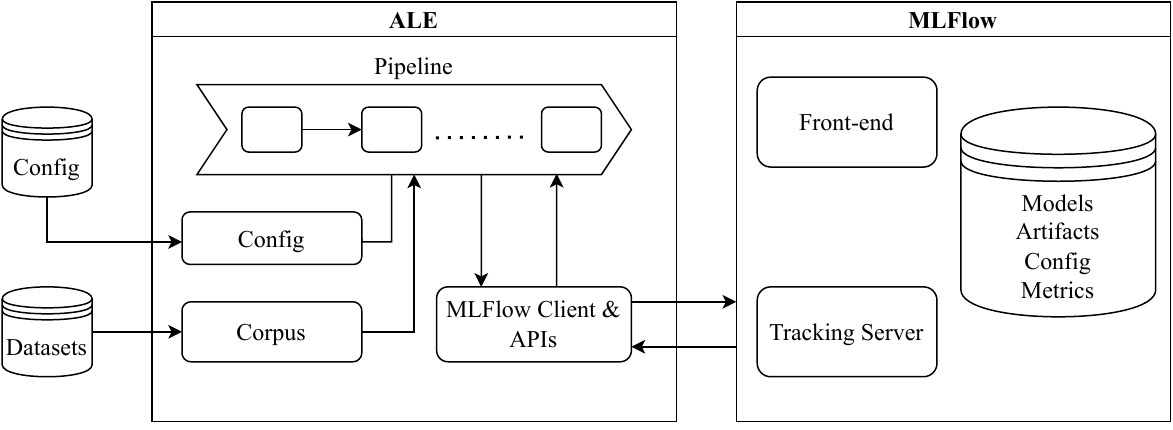}
    \caption[Architecture Overview]{The architecture consists of two components: \textit{ALE} itself and \textit{MLFlow}. We provide both components as dockerized applications. Every pipeline step will log its configuration, metrics, and artifacts to the tracking server. The data scientist uses \mlflow to inspect and compare experiments via the \mlflow front-end.}
    \label{fig:aletecture}
\end{figure}

The architecture of our python framework focuses on flexibility and ease of adoption by leveraging open-source tools and allowing developers to integrate their own workflows and machine learning libraries. We use \mlflow to track configurations, pipeline steps, artifacts, and metrics. Dockerization allows users to execute the framework remotely. \autoref{fig:aletecture} shows a brief overview of our architecture.

\subsection{Configuration}
\label{sec:configuration}
The configuration system is one of the core components of the framework. It allows setting up different environments for testing \alstrats. The hydra framework \cite{yadanHydraFrameworkElegantly2019} achieves this by dynamically creating a hierarchical configuration. This configuration is composed of different files and command line parameters. To set up a reliable end-to-end pipeline to test different strategies, we provide the following configuration modules: \textit{converter}, \textit{data}, \textit{experiment}, \textit{teacher}, \textit{trainer}, and \textit{MLFlow}. See GitHub for detailed explanations.

We provide a simple way to create reproducible experiments: We leverage the python API of \mlflow and combine it with hydra to make use of \mlflows tracking capabilities and hydra's powerful configuration system.
The configuration is stored in \textit{.yaml} files. When running the framework, the configuration files are processed by
hydra, stored as python objects and accessible to the pipeline components (\autoref{sec:Pipeline}). The configuration object is passed down and logged to the \mlflow tracking server. The logged configuration enables the framework to detect already processed pipeline steps and load
them from the tracking server, providing a simple yet powerful caching mechanism.

\subsection{Dataset}
\label{sec:dataset}
NLP tasks like text classification and span labeling are based on big amounts of labeled data. To focus on the comparison of different \alstrats and not on converting data, we adopt the \textit{convention over configuration} principle \cite{baechleRubyRails2007} to \textit{convention over coding}. If the users follow these conventions, they do not have to write code, but at most, a short configuration. 

We introduce the following conventions:
(1) The raw data should be in JSONL format. (2) The converted format is \spacy DocBin. (3) We add ids to each raw entry and use it for global identification throughout the framework.

At the moment, the framework handles text classification and span labeling tasks. In the case of span labeling tasks, each data point contains a list of the labeled spans: each span is represented by the start character offset, end character offset, and the according label (\verb|"labels":[[0,6,"LOC"]]|). For text classification tasks, the label is simply the label name (\verb|"label":"neg"|)\footnote{In multi-label classification tasks the label is a list of labels.}. For our case study, we used the IMDb corpus \cite{maasLearningWordVectors} and the dataset originating from the Text RETrieval Conference (TREC) \cite{hovySemanticsBasedAnswerPinpointing2001}. \autoref{sub:case-data} shows details.

\subsection{Pipeline}
\label{sec:Pipeline}
\begin{figure}[tb]
    \centering
    \includegraphics[width=0.9\linewidth]{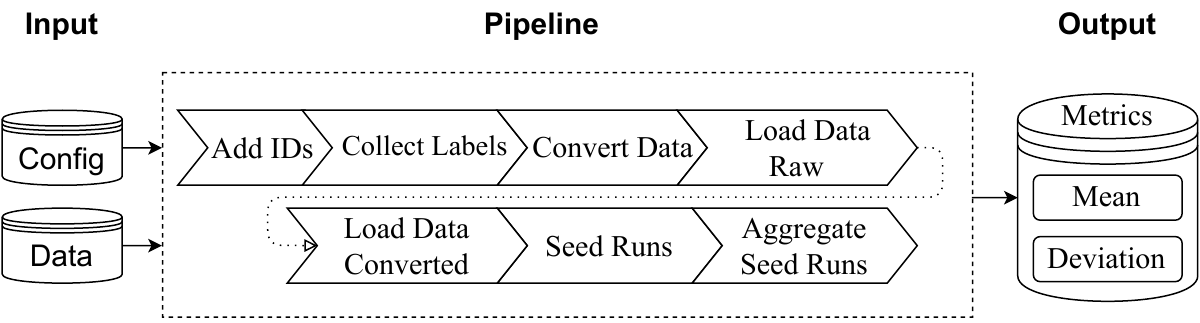}
    \caption[Pipeline Overview]{ALE uses the shown pipeline architecture for the experiments. The \textit{Load
    Data Raw} and \textit{Load Data Converted} will copy the input data to the working directory and rename the files according to the convention. The copied data is logged to the tracking server. Tracking all the input and metadata allows running subsequent runs remotely, fetching data from the tracking server as needed. \autoref{fig:seed_runs} shows details for \textit{Seed Runs} and \textit{Aggregate Seed Runs}.}
    \label{fig:pipeline}
\end{figure}

The framework provides a customizable yet fixed setup. We decided to build a step-by-step pipeline with customizable components. The \textit{MLFlowPipeline} class acts as the executor. It collects pipeline components and initializes the pipeline storage and configuration. The executor will run every child pipeline component after another. 
We use \mlflow to track our experiments. Each pipeline step reports parameters and results to their \mlflow \textit{run}. \autoref{fig:pipeline} shows the pipeline steps. Order of execution is essential as the different components depend on each other. For example, the \textit{load data converted} step needs the data from the \textit{convert data} step. Before the framework executes a pipeline step, it will check if a run with matching parameters and revision was already executed. To achieve this functionality, the executor and every component log their run parameters to MLFlow. If a run matches and the match is marked as successful, the framework fetches the result of the run. Otherwise, if the matched run is marked as failed, the run will try to resume the failed run.
\begin{figure}[htb]
    \centering
    \includegraphics[width=0.75\linewidth]{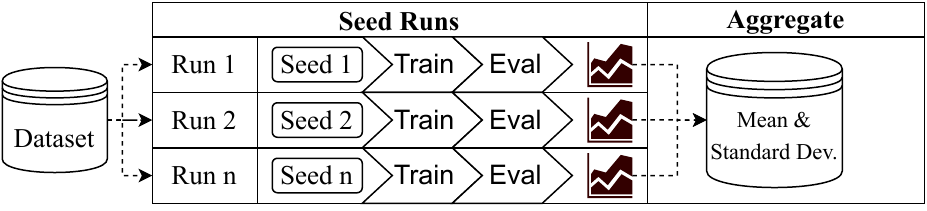}
    \caption[Aggregating the AL simulation with different seeds]{The last two steps of our pipeline contain the actual core logic. The \textit{Seed Runs} component step starts n-different simulation runs, each with its own distinct seed. A run iteratively proposes new data based on the strategy and trains and evaluates the model. This will lead to n-different results for one strategy. To get one comparable result, the \textit{Aggregate Seed Runs} step will fetch the resulting metrics and calculate the mean and deviation over all runs.}
    \label{fig:seed_runs}
\end{figure}

\section{Core Composition}
\label{sec:core-composition}
\begin{figure}[htb]
    \centering
    \includegraphics[width=0.8\linewidth]{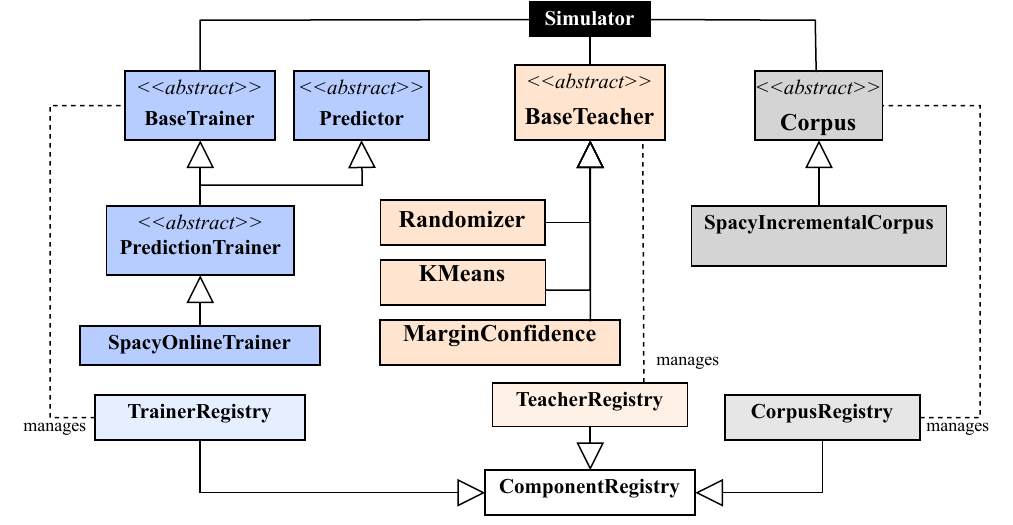}
    \caption{The \textit{simulator} (black) holds instances of the \textit{trainer} (blue), \textit{teacher} (beige), and \textit{corpus} (grey). The configuration specifies which implementations the simulator fetches from the registries. The registry knows all implementations of each component. We distinguish between \textit{BaseTrainer} and \textit{Predictor} to hide the training API for the teachers. The teacher can only perform predictions but no training. }
    \label{fig:UML_bartender}
\end{figure}

Four components build the core unit of the framework: \textit{simulator}, \textit{teacher}, \textit{trainer}, and \textit{corpus} (see \autoref{fig:UML_bartender}). These form a logical unit responsible for selecting data proposed to the simulated oracle using gold labels as annotations (see \autoref{fig:SimulationCycle}). The \textit{registry} classes collect all implementations of the abstract classes \textit{teacher}, \textit{trainer}, and \textit{corpus} according to an adaption of the behavioral registry pattern\footnote{\url{https://github.com/faif/python-patterns/blob/master/patterns/behavioral/registry.py}} used in Python.

The \textbf{simulator} class\footnote{The simulator is called \textit{AleBartender} and \textit{AleBartenderPerSeed} in the code.} takes care of the high-level management of the data proposal by setting up the instances of corpus, trainer, and teacher using the classes provided by the configuration. Hereby, the simulator runs active learning cycle simulation (see \autoref{fig:ALCycle}) for every seed as a run on MLFlow.
A \textbf{corpus} instance tracks the annotation progress: e.g., which documents the simulator has already annotated and which documents need annotations. Furthermore, it provides a trainable corpus based on the annotated data. Thus, the corpus and trainer represent a unit.
The \textbf{trainer} uses a ML/DL framework to train and evaluate a model. We provide a default implementation \textit{SpacyOnlineTrainer}, with \spacy (see \autoref{sec:trainer}). We allow the teachers to make predictions with the trained model. Thus, the trainer class implements the abstract methods of the \textit{Predictor}.
The \textbf{teacher} realizes the selection of unlabeled potential data points for the annotation simulation (see \autoref{sec:teacher}).

The structure of this unit allows users to implement their own components, such as \alstrats the framework uses for evaluation: the user has to implement the abstract class. If the strategy follows the folder convention, the registry finds the implementation, and the user can evaluate the new strategy via the hydra configuration. The same applies to the other components.

\subsection{Teacher}
\label{sec:teacher}
The basic functionality of the teacher component is to propose new data points to the \textit{Simulator} (see \autoref{fig:UML_bartender}). Therefore, a teacher needs to implement the \textit{BaseTeacher's} abstract method presented in \autoref{listing:propose}.

Additionally, if needed, the teacher may use metrics from training by implementing the BaseTeacher's \textit{after\_train()} and \textit{after\_initial\_train()} methods to perform further computations after the corresponding training step, depending on the implemented AL 
strategy\footnote{For instance, multi-armed bandit strategies use the gain in model performance after each training to estimate the reward over clusters of data to draw from \cite{agrawalAnalysisThompsonSampling2012,kuleshovAlgorithmsMultiarmedBandit2014}.}.
AL teacher strategies are broadly divided into two areas, namely exploration and exploitation (see \autoref{sec:related-work}). 
While purely exploration-based strategies are model-independent, exploitation-based strategies exploit the model's state of training and, thus, are model dependent \cite{loyStreambasedJointExplorationexploitation2012,bonduExplorationVsExploitation2010}.
ALE is required to handle both classes of AL strategies and hybrid techniques that use exploitation and exploration. Therefore, the teacher takes the \textit{predictor} as a parameter (see \autoref{fig:UML_bartender}). 

\begin{figure}[htb]
    \begin{lstlisting}[caption={BaseTeacher's abstract propose \textit{method} to be implemented. The \textit{propose} method takes a list of potential (unlabeled) data point ids, the actual step size, and the budget of the propose step. Step size and budget are limited to the remaining length of the not annotated data points. The method returns a list of data point ids according to the corresponding AL strategy.}, label={listing:propose}, basicstyle=\fontsize{8}{8}\selectfont\ttfamily]
@abstractmethod
def propose(self, 
            potential_ids: List[int], 
            actual_step_size: int, 
            actual_budget: int) -> List[int]:
    # See text for documentation
    pass
\end{lstlisting}
\end{figure}

\subsection{Trainer}
\label{sec:trainer}
The trainer component takes the proposed data in each iteration and trains a model. The trainer first evaluates the model on the development dataset to pass information to the teacher. Afterward, it evaluates the model on the test set for performance comparison. Thus, the framework provides feedback on the quality of the newly proposed batch based on the objective metric, which serves as a benchmark for comparing the performance of different AL strategies. 

ALE offers an abstraction layer that enables users to use an arbitrary machine learning or deep learning framework, such as PyTorch, TensorFlow, or PyTorch Lightning. Therefore, the user must implement the abstract methods specified in the \textit{PredictionTrainer} class: 
\textit{train}, \textit{evaluate}, \textit{store/restore} for the resume process, and \textit{predict} for exploitation strategies.

We focus on the NLP domain, and thus we select \spacy as the baseline framework due to its sophisticated default implementations and practicality for production use cases \cite{montaniExplosionSpaCy2023}. During each iteration of the active learning process, the framework generates a new batch of data to be annotated and calls the trainer with the updated corpus.

The trainer initiates the \spacy training pipeline in the constructor and reuses it in each subsequent training iteration. 
We also tested training from scratch after each \textit{propose step}. However, this approach takes time to create the \spacy pipeline\footnote{We empirically tested different configurations and noticed computation time reduction up to ten percent.}. Thus, we switched to online/continual learning \cite{masanaClassincrementalLearningSurvey2022,hoiOnlineLearningComprehensive2018}. Continual learning has other challenges, like the \textit{catastrophic forgetting problem} \cite{kaushikUnderstandingCatastrophicForgetting2021,kirkpatrickOvercomingCatastrophicForgetting2017}.
The framework addresses this issue by training not only on the new proposed data but also on all previously proposed data. 
Thus, the trainer starts with a checkpoint from the previous iterations and fine-tunes the weights with all data available up to that iteration\footnote{We compared the average of five simulation runs with and without the online approach. The latter achieves faster convergence. The final scores are nearly equal.}.

\section{Case Study}
\label{sec:case-study}
To demonstrate the implementation of different datasets and \alstrats in ALE, we implemented an exploitation and an exploration based AL strategy. Both strategies are pool-based and applied to text classification tasks. We evaluated their performance compared to a randomizer and visualized the results in ALE on the IMDb sentiment dataset \cite{maasLearningWordVectors} and the TREC dataset \cite{liLearningQuestionClassifiers2002,hovySemanticsBasedAnswerPinpointing2001}.
As we stick to the convention of using a \spacy model, we may simply use the implemented pipeline components. Therefore, we use the \textit{SpacyIncrementalCorpus} implementation of ALE's \textit{Corpus} class for both corpora and the implemented \textit{SpacyOnlineTrainer}. 

\subsection{Experiment Setup}
\label{sub:case-data}
We compared the strategies on two text classification tasks: a binary sentiment classification task with the IMDb reviews corpus and a multiclass text classification task with the TREC coarse dataset. 

The \textit{IMDb Reviews Corpus} is a dataset for binary sentiment classification \cite{maasLearningWordVectors}. As a benchmark dataset, it provides $50,000$ documents, split into a training set of size $25,000$ and a test set of size $25,000$, which we randomly split into a test and dev set of sizes $12,500$ each.

\textit{Originating from the Text RETrieval Conference (TREC)} is a multiclass question classification dataset\cite{hovySemanticsBasedAnswerPinpointing2001}. It contains $5,452$ annotated questions for training and $500$ annotated questions for testing. For our case study, we further split the training set to train and dev set, s.t. we obtained a total of $546$ annotated documents (10\%) for evaluation and $4906$ documents for training. 
We trained our model on the coarse labels for our case study, consisting of five classes.

The two presented datasets differ in complexity. The sentiment analysis task involves larger documents and a comparatively simple binary classification task. In contrast, the question classification task consists of short documents (single sentences/questions) and a comparatively difficult multiclass classification task with five classes.   
For our case study, we use SpaCy's bag-of-words \textit{n-gram model} for sentiment classification\footnote{For details see \url{https://spacy.io/api/architectures\#TextCatBOW}} for runtime efficiency and as we considered it sufficient for the merely demonstrational purposes of the case study. However, on the question classification task, we did not obtain usable results with the simple model and, thus, used SpaCy's \textit{transformer model} for question classification\footnote{For details see \url{https://spacy.io/api/architectures\#transformers}}.

\subsection{Implemented Active Learning Strategies}
We implemented and compared an exploration- and an exploitation-based \alstrat for the case study, as the two classes of \alstrats are considered to form a trade-off. 
While exploration-based methods are computationally less expensive and generally yield a more representative sample of the data such that they increase the external validity of the model, exploitation-based methods have shown significant theoretical and practical performance improvements for many domains \cite{zhangExplorationExploitationAdaptive2003,bonduExplorationVsExploitation2010}. 
For comparison, we implemented a baseline randomizer strategy. The randomizer draws a random sample of step size $N$ data points for each propose step. Therefore, we needed to implement the \textit{propose} method of the \textit{BaseTeacher}. 

\textit{Exploration-Based Teacher (K-Means).}
The implemented exploration-based \alstrat is based on the cluster hypothesis \cite{kurlandClusterHypothesisInformation2014}. The cluster hypothesis claims that documents within each cluster are more likely to belong to the same class, i.e., to have the same prospective label. 
The teacher initially clusters the datasets using TFIDF-vectors and the k-means algorithm \cite{shahDocumentClusteringDetailed2012}, where $k$ equals the number of labels. The further the document is away from the center of its respective cluster, the closer it is assumed to be to the decision boundary \cite{ganKmeansBasedActive2017}. Thus, given a step size $N$, the teacher returns the data points farthest from the respective centers in each propose step. To do so, we need to implement the \textit{propose} method of the teacher only. The propose method returns the farthest $N$ unseen documents from the centers according to the euclidean distance\footnote{We follow \cite{ganKmeansBasedActive2017} using the euclidean distance and TFIDF-vectors. Other approaches such as cosine similarity and doc2vec are interesting alternatives.}. There is no need to implement \textit{after\_train} or the \textit{after\_initial\_train} methods of the BaseTeacher (see \autoref{sec:teacher}), as the presented \alstrat does not take any feedback from the training.

\textit{Exploitation-Based Teacher (Margin-Confidence).}
For exploitation, we implemented a margin-confidence AL strategy. Margin-based \alstrats assume that, given a data point, smaller margins between the possible prediction classes of the model mean less certain predictions by the model \cite{joshiMulticlassActiveLearning2009}. For instance, we may assume a sentiment classifier to be more confident if the prediction \textit{negative} has a considerably higher confidence value than \textit{positive}, compared to both possible predictions having almost equal confidence values.
Thus, for each propose step, the teacher needs the current state of the trained model to make predictions on the remaining unlabeled data. To reduce the computational costs, the implemented teacher only predicts a random sample of size $M$, which denotes the \textit{budget} of the \alstratpredot. 
For the binary case, the margin confidence for a given data point is given by the margin between the scores of both labels.
In the multiclass case, we adopted the best versus second best label, as stated in the margin-confidence approach by \cite{joshiMulticlassActiveLearning2009}. The margin between both classes with the highest predicted scores constitutes the margin confidence of the model for a given data point. Then, the teacher returns the step size $N$ data points with the smallest margin confidences. 
As for the exploration-based teacher, there is no need to implement the BaseTeacher's \textit{after\_train} or the \textit{after\_initial\_train} methods.

\subsection{Experiment Configuration}
\label{sec:exp-conf}
The margin-confidence teacher uses a budget of 5000 datapoints. The experiment is configured for the case study to use a GPU, a step size of $N_1 = 1000$, and a second step size of $N_2 = 0.2*M$, where $M$ is the  dataset size. The initial training ratio is $0.05$ for all experiments conducted. The tracked metric is the macro F1 score \cite{fawcettIntroductionROCAnalysis2006a}. In principle, any other performance metric could be used at this point if correctly implemented by the respective trainer. Especially if the \textit{after\_train} methods are implemented, the teacher receives the full metrics set implemented by the trainer and, thus, may use different metrics for proposing than the tracking metrics.
Finally, both \alstrats and the randomizer were evaluated on $5$ random seeds ($42$, $4711$, $768$, $4656$, $32213$).

Experiments were conducted using a private infrastructure, which has a carbon efficiency of 0.488 kgCO$_2$eq/kWh. A cumulative of 50 hours of computation was performed on hardware of type RTX 8000 (TDP of 260W).
Total emissions are estimated to be 5.62 kgCO$_2$eq of which 0 percents were directly offset.
Estimations were conducted using the \href{https://mlco2.github.io/impact#compute}{MachineLearning Impact calculator} presented in \cite{lacosteQuantifyingCarbonEmissions2019}.

\subsection{Results}
We evaluated both \alstrats compared to the baseline randomizer on all 5 seeds, as given in \autoref{sec:exp-conf}.

\begin{figure}[htb]
    \centering
    \begin{subfigure}{0.49\textwidth}
            \includegraphics[width=\textwidth]{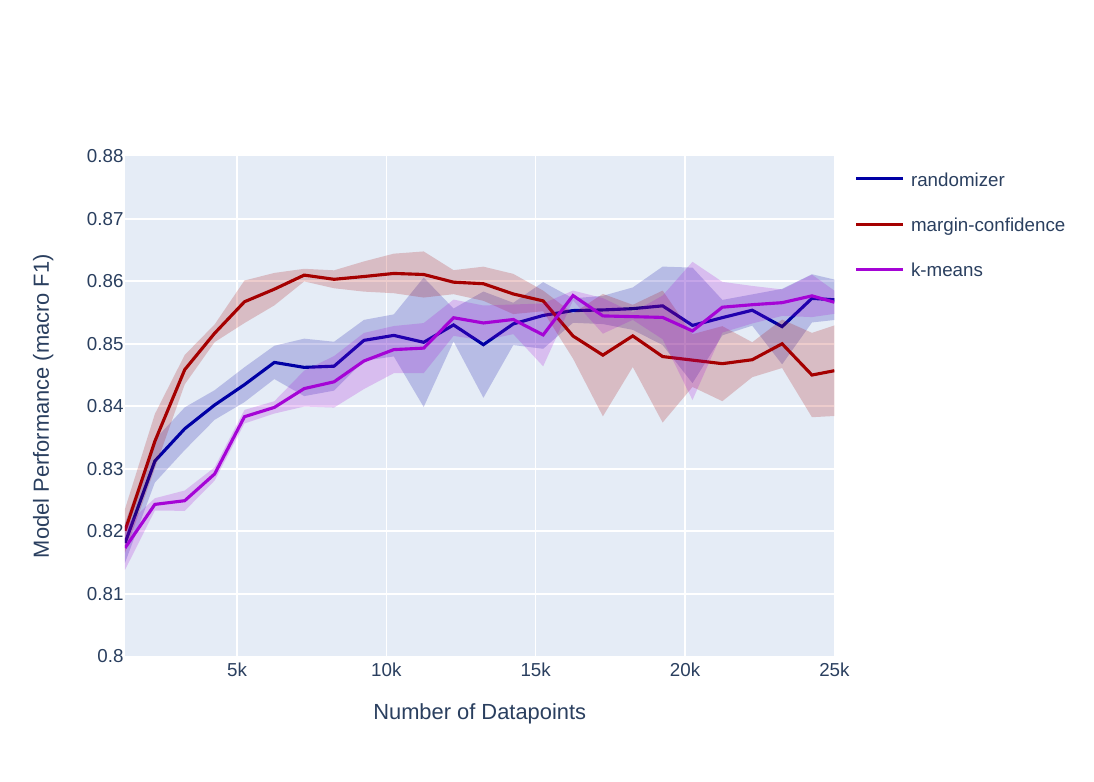}
            \caption{step\_size $N_1 = 1000$}
            \label{fig:res_imdb_1000}
    \end{subfigure}
    \begin{subfigure}{0.49\textwidth}
            \includegraphics[width=\textwidth]{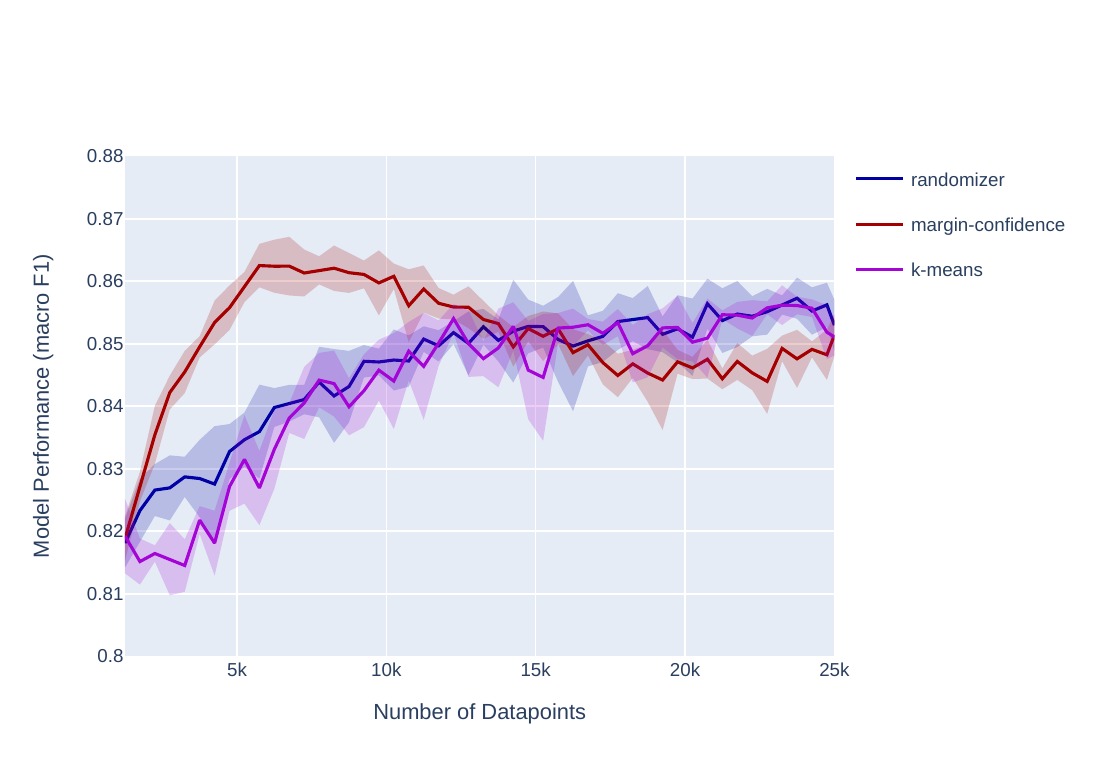}
            \caption{step\_size  $ N_2 = 500$}
            \label{fig:res_imdb_500}
    \end{subfigure}
    \begin{subfigure}{0.49\textwidth}
            \includegraphics[width=\textwidth]{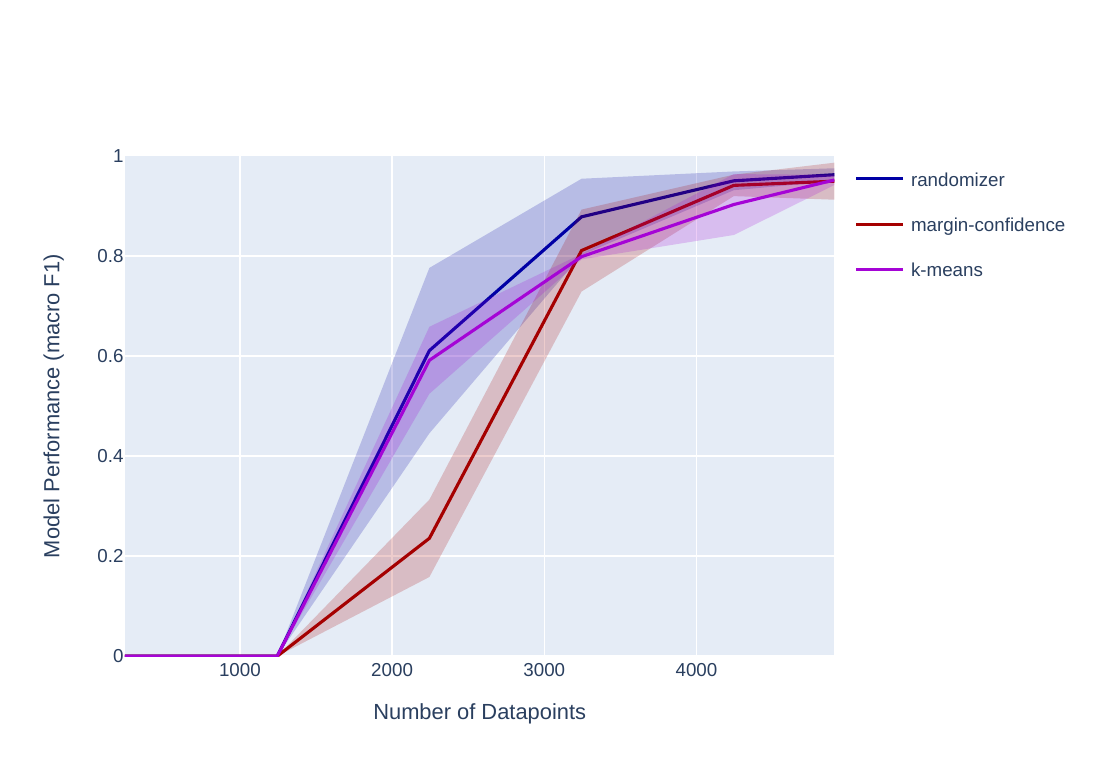}
            \caption{step\_size $N_1 = 1000$}
            \label{subfig:trec1000}
    \end{subfigure}
    \begin{subfigure}{0.49\textwidth}
            \includegraphics[width=\textwidth]{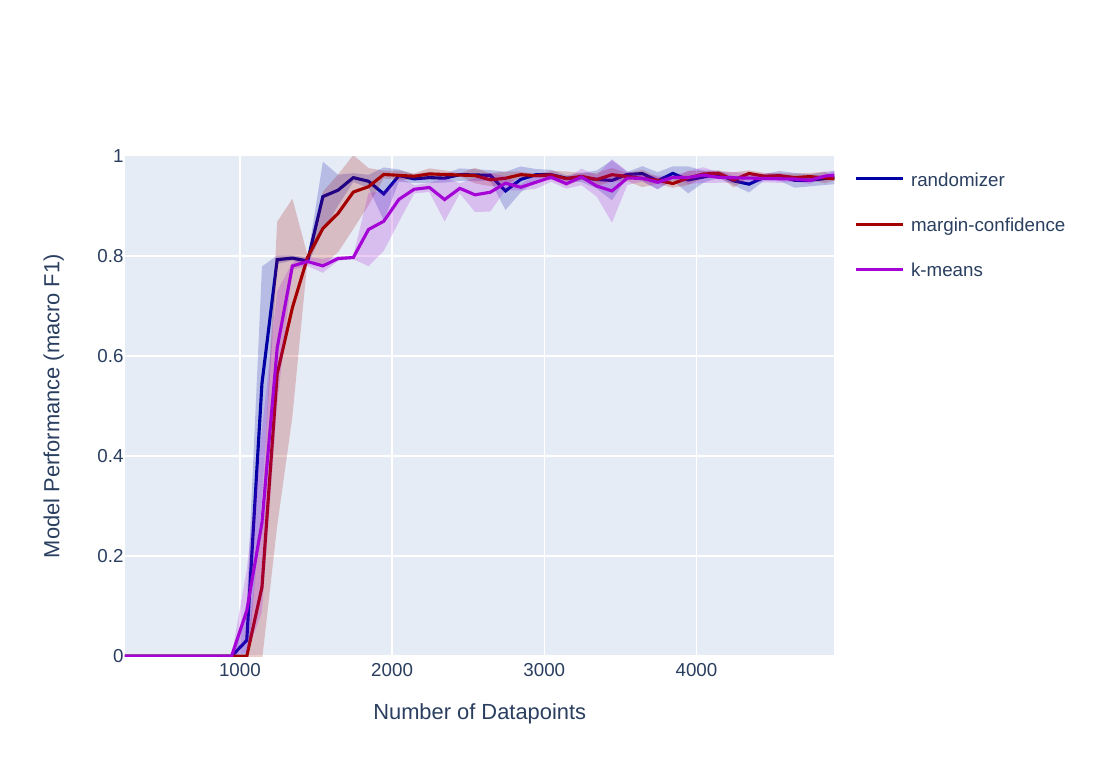}
            \caption{step\_size $N_2 = 100$}
            \label{subfig:trec100}
    \end{subfigure}
    \caption{Mean macro F1-scores of the models for each propose step on the IMDb dataset (a, b) and the TREC Coarse dataset (c, d) with deviations (min-max).}
    \label{fig:results}
\end{figure}

\textit{IMDb Sentiment Classification.}
On the IMDb reviews dataset, the margin-confidence strategy performed superior to both the randomizer and the k-means strategy for $N_1$ and $N_2$.
For step size $N_1$, the k-means strategy performed inferior to the randomizer. The margin-confidence strategy reaches the threshold of $f1 = 0.85$ on average after proposing $5250$ data points only. The randomizer reaches that threshold after proposing $9250$, and the k-means teacher after proposing $12250$ data points (see \autoref{fig:res_imdb_1000}).
Similarly, for step size $N_2$, we observed a slightly inferior performance of the k-means strategy compared to the randomizer. The randomizer reaches the threshold of $f1 = 0.85$ after proposing $11250$ data points, while the k-means strategy reaches the threshold after $11750$ data points. Again, the margin-confidence strategy outperformed both the randomizer and the k-means strategy by a significant margin, reaching the threshold after proposing $4250$ data points (see \autoref{fig:res_imdb_500}).

\textit{TREC Question Classification.}
On the coarse question classification task of the TREC dataset, we observe a minimum threshold of $\approx 1000$ proposed data points for the transformer model to learn from the data. With this insight, the initial data ratio may be adapted. For $N_1$, both the k-means and the margin-confidence strategy performed inferior to the randomizer. All three \alstrats exceeded the threshold of $f1 = 0.85$ after proposing $4245$ data points. However, the average $f1$ score of the randomizer was higher than the scores of the k-means and the margin-based strategy for every propose step. Yet, the randomizer showed the highest seed dependent deviation (see \autoref{subfig:trec1000}). Therefore, the k-means strategy could, given its low seed-dependent deviation and comparatively good performance, be a more stable strategy for annotating data points for this task.

For $N_2$, we also do not observe significant advances in the \alstrats compared to the randomizer. The randomizer performs superior to the k-means strategy and the margin-based strategy. More specifically, while both the randomizer and the margin-confidence strategy reach the threshold of $f1=0.85$ after proposing $1545$ data points, and the k-means strategy after proposing $1845$ data points (see \autoref{subfig:trec100}), the randomizer's scores are ahead of the magin-confidence strategy's scores for most propose steps.

\subsection{Discussion}
This case study does not represent a scientific investigation on the performance of the presented strategies but demonstrates the purposes and the usability of the framework presented in this paper.
The k-means strategy performs poorly in our case study and should be investigated further. A possible source of its performance may be the vectorization or the within-cluster homogeneity. If a single cluster has larger within-cluster distances, it may be proposed disproportionately often, thus introducing biases to the proposed dataset.
Furthermore, the strategies were tested for different datasets and different models. Comparing their performance on multiple datasets using the same model could reveal further insights.
Both strategies did not perform significantly better than the randomizer for the TREC coarse task, neither for $N_1$ nor $N_2$. Further experiments with, first, more random seeds to increase generalizability and, second, other classifiers may gain further insights on this matter. Moreover, the initial data ratio should be adopted. 
Lastly, we observed a task-dependent effect of step sizes on the model's overall performance. Given the lower step size $N_2$, the trained models exceeded the given threshold earlier and reached significantly better results after proposing significantly fewer data points (e.g., $f1>0.9$ after $\approx 1545$ data points for $N_1$, after $\approx 4245$ data points for $N_2$). The chosen model architecture and the online training convention may cause this effect. Still, the step size can be identified as one of the relevant parameters to investigate when evaluating \alstrats using ALE. 
On the IMDb sentiment analysis task, we observed the margin-confidence strategy after training on fewer data points to perform better than the randomizer after all data points. A possible reason may, again, be the warm start of the model's training, being initialized with the previous model's weights and, thus, running into a different local optimum.
A scientific evaluation of the effect would be desirable.

\section{Conclusion and Future Work}
\label{sec:conclusion}
This study presents a configurable and flexible framework for the fair comparative evaluation of AL strategies (\textit{queries}). The framework facilitates the implementation and testing of new AL strategies and provides a standardized environment for comparing their performance. It tracks and documents all parameters
and results in a central MLFlow instance. Due to the long-running active learning simulation computations, we provide containerization for using cloud infrastructure and parallelize independent computations. The demonstration of two AL strategies and the instructions for implementing additional strategies shows the usage and may motivate other researchers to contribute their query strategies.

We plan to conduct a survey for developing best practices in the field, enabling researchers and practitioners to select appropriate strategies for their specific NLP task reliably.
Additionally, an expansion to the area of relation extraction tasks would allow us to reduce annotation effort in a highly time-consuming domain referring to the annotation process in the NLP domain.

\bibliographystyle{splncs04}
\bibliography{AL-Group}

\end{document}